\documentclass[runningheads]{llncs}
\usepackage{graphicx}
\usepackage{amsmath,amssymb} 
\usepackage{color}
\usepackage[width=122mm,left=12mm,paperwidth=146mm,height=193mm,top=12mm,paperheight=217mm]{geometry}
\usepackage{placeins}
\usepackage{times}
\usepackage{epsfig}
\usepackage{mathrsfs}
\usepackage{bm}
\usepackage{subcaption}
\usepackage{hyperref}
\usepackage{url}
\usepackage{cite}

\begin{document}

\pagestyle{headings}

\mainmatter

\title{\textsc{Line Artist}: A Multi-style Sketch to Painting Synthesis Scheme} 




\author{Jinning Li \and Siqi Liu \and Mengyao Cao}
\institute{Department of Computer Science and Engineering,\\
Shanghai Jiao Tong University\\
\texttt{\{lijinning, magi-yunan, cmy\_0622\}@sjtu.edu.cn}}
\maketitle

\begin{abstract}
Drawing a beautiful painting is a dream of many people since childhood. In this paper, we propose a novel scheme, \textsc{Line Artist}, to synthesize artistic style paintings with freehand sketch images, leveraging the power of deep learning and advanced algorithms. Our scheme includes three models. The \textit{Sketch Image Extraction} (SIE) model is applied to generate the training data. It includes smoothing reality images and pencil sketch extraction. The \textit{Detailed Image Synthesis} (DIS) model trains a conditional generative adversarial network to generate detailed real-world information. The \textit{Adaptively Weighted Artistic Style Transfer} (AWAST) model is capable to combine multiple style images with a content with the VGG19 network and PageRank algorithm. The appealing artistic images are then generated by optimization iterations. Experiments are operated on the Kaggle Cats dataset and The Oxford Buildings Dataset. Our synthesis results are proved to be artistic, beautiful and robust.

\end{abstract}

\begin{center}
    \centering
    \includegraphics[width= 1\textwidth]{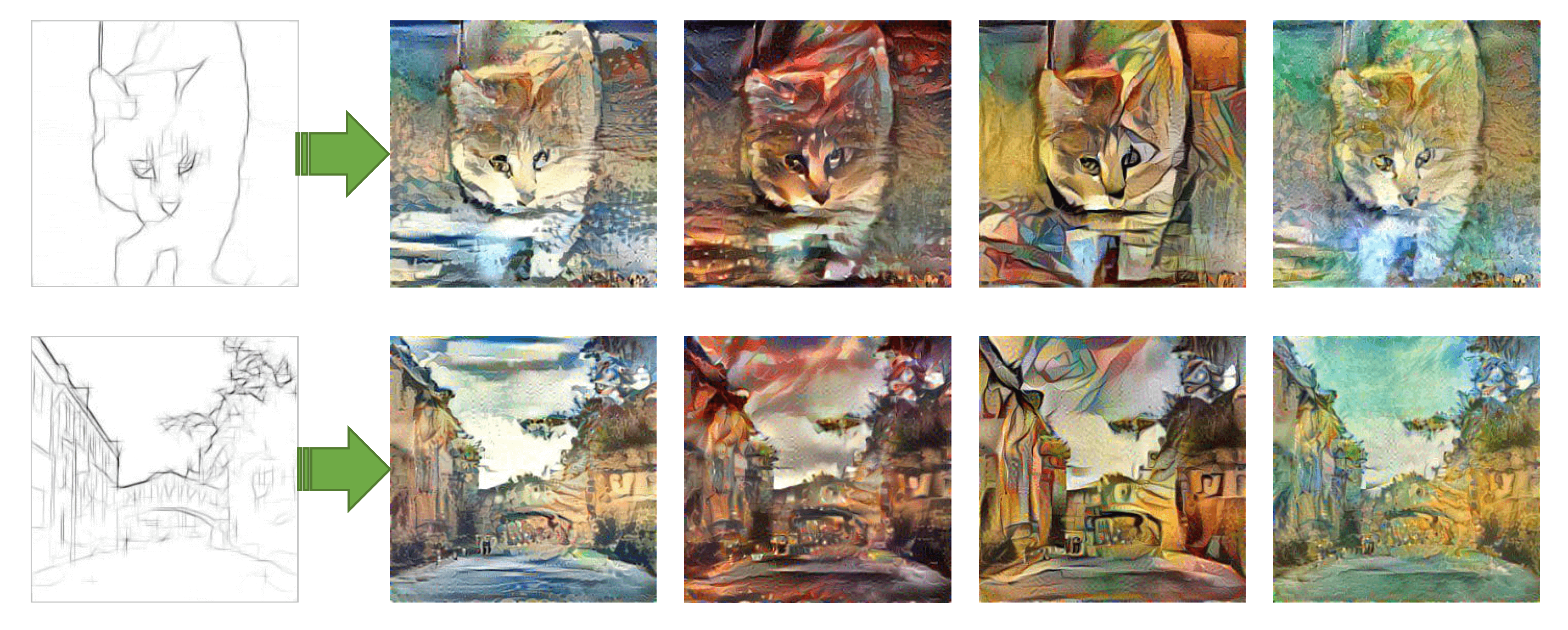}
    \captionof{figure}{Artistic paintings synthesized by \textsc{Line Artist}. Left: inputted freehand sketch images. Right: synthesized artistic paintings of different styles}
    \label{fig::heading}
\end{center}%

\section{Introduction}
Deep neural networks and Generative adversarial networks (GANs) has been applied to many scientific and engineering fields. However, researches on artist production are still limited. There are many people who want to paint beautiful paintings but feel depressed about being not good at drawing and coloring. So we are considering creating a method using machine learning to assist people in drawing artistic paintings.

In this paper, we propose a new scheme, \textsc{Line Artist}, to paint like a well-known painter. The only thing needed to be done is to draw some sketch lines. Then, \textsc{Line Artist} will draw a meaningful and elegant painting for the users like images shown in Fig.\ref{fig::heading}. Everybody can become a skilful painter!

Many works have been done to transform a real-world image into an artistic one. However, in our work, the inputs are sketch images containing much less information instead real photos because we aim at building a more convenient artistic style generating system for the users. This task is much harder than others since we need to obtain more information with limited information inputted.
\begin{figure}[!ht]
  \centering
  \includegraphics[width = 1\linewidth]{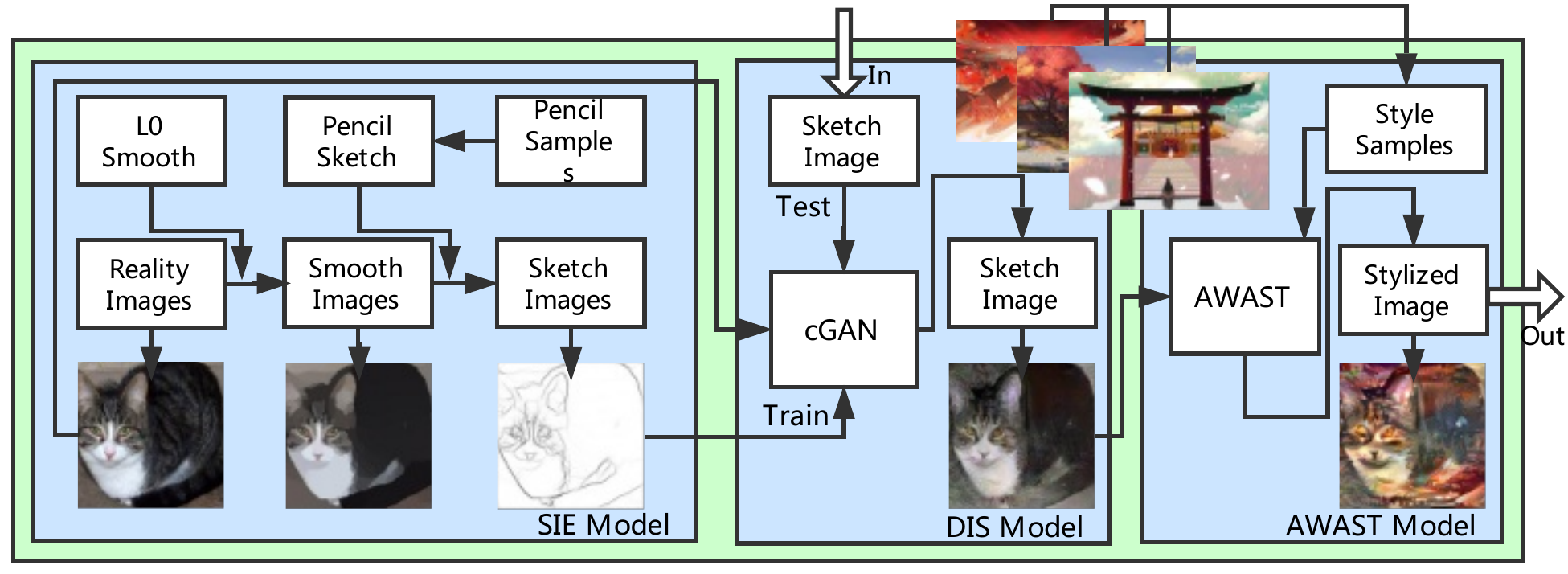}
  \caption{Overview of \textsc{Line Artist}. Left: SIE model used to generate the dataset for training. Middle: DIS model based on cGAN used to add informative details into sketches. Right: AWAST model for stylizing images with multiple style samples.}
  \label{fig::overallFlow}
\end{figure}

In order to generate more information. Our scheme employs the GANs to extract the information from the sketch drafts and generate detailed images with more information, which is shown on the middile of Fig.\ref{fig::overallFlow}. Then, we propose a novel multi-style transfer algorithm based on the Artistic Style Transfer~\cite{gatys2015neural} algorithm and PageRank~\cite{page1999pagerank} to transform our synthesized detailed images into artistic ones.

The first challenge is the dataset, there are few usable datasets about the sketch images and their corresponding real-world images. However, in order to train a supervised system that is capable of generating a more informative detailed image using a sketch image, datasets are necessary. Inviting volunteers to tag a new dataset will cost a large scale of time and human resources. To solve this problem, we introduce the \textit{Sketch Image Extraction} (SIE) model to synthesize sketch-like images and build the dataset efficiently, which is very similar to real freehand sketches. The SIE model is shown on the left side of Fig.\ref{fig::overallFlow}.

To extract the sketch more accurately in the SIE model, we apply the $L_0$-Smooth~\cite{xu2011image} algorithm to smooth the initial image. This process will make the edges more distinct and wipe out the unnecessary veins. Then, we adapt the pencil drawing~\cite{lu2012combining} method to extract the sketch images from the smooth ones. Through the SIE model, we obtain the dataset containing pairs of the generated sketch images and their corresponding real-world images.

To generate informatively detailed images from given sketch images, the \textit{Detailed Image Synthesis} (DIS) model is introduced whose procedure is shown on the middile of Fig.\ref{fig::overallFlow}. We use the dataset generated by the SIE model to train a system that receive the skech images extracted and output detailed images by generating more information. In this paper, we adapt the Pix2Pix~\cite{isola2016image} model with a new generator to the realize this synthesis operation. Experiments prove that the DIS model can also synthesize a nice result even though with the dataset generated by computer in the SIE section.

Finally, the \textit{Adaptively Weighted Artistic Style Transfer} (AWAST) model, shown on the right of Fig.\ref{fig::overallFlow}, solves the challenge to transfer the style of images with multiple painting samples for a selected style. Artist Style Transfer algorithm proposed in~\cite{gatys2015neural} is able to combine two images with both low-level features and high-level features extracted leveraging a pretrained VGG19 network~\cite{simonyan2014very}. In this paper, we introduce the \textit{Adaptive Style Weight} (ASW) based on the feature similarity network and PageRank algorithm. Using ASW, multiple painting images can be combined more naturally and beautifully with the content. After these process, a colorful painting will be obtained using just a line sketch drawn by the users.

The contributions of this paper are summarized as bellow:
\begin{itemize}
  \item A delicate sketch image extracting scheme and two elaborate datasets containing pairs of real-world images and their corresponding sketch images.
  \item An efficient detailed image synthesis model achieving more real-world details and patterns by inputing sketch images.
  \item A novel algorithm to adaptively combine multiple painting samples with the content and synthesize appealing artistic images.
\end{itemize}

\section{Related Works}
\paragraph{\textbf{Sketch Extraction}}
Many researches~\cite{gonzalez2016improved, kumar2017fractional, gao2010improved} focus on the edge extraction based on algebraic algorithm like sobel operator and fuzzy mathematics. These traditional edge extraction methods are good substitutions of sketch and usually run fast. However, the trends and continuity of extracted edges are not as natural as man-made ones. In~\cite{Dollar_2013_ICCV}, the author propose a random forest based method to detect edges. In the reference~\cite{xie15hed}, the author propose a CNN-based edge detection algorithm, which performs better than the traditional ones.

In the paper~\cite{wei2017road}, the author propose a RSRCNN to extract roads from aerial images, which can also be applied to the sketch extraction. However, the CNN-based methods are highly relied on the training datasets and cost a lot of resources to train a network. In~\cite{zhang2015face} the author propose a face sketch synthesis scheme base on greedy search, this technique can synthesis sketch for other objects. However, its effect is more like a grey transfer than sketch extraction.
In~\cite{zhang2015segment, xu2011image}, efficient methods of image smoothing are proposed based on wave pattern, which is helpful preprocessing for extraction.
In~\cite{lu2012combining}, Lu propose a fast scheme to synthesize pencil drawing sketch image. The result is very appealing.

\textbf{Detail Synthesis}
In~\cite{sangkloy2016scribbler, isola2016image, odena2016conditional}, methods extended from GAN are used to synthesize detailed images with more information from given materials. These models are usually easier to train while the result are more fuzzy to some extent. Chen proposed an end-to-end cascaded refinement networks in~\cite{chen2017photographic} to synthesize large-size reality image from semantic layouts, whose result has a high resolution and accuracy. But this method is highly dependent on training datasets.
In~\cite{ho2016laplacian}, the author propose an image synthesis model based on Laplacian pyramid, which has a lower computation complexity.
In~\cite{pathak2016context}, Context Encoders,  a GAN based model, is promoted to generate more information from the surroundings, in which artist style could be applied.

\textbf{Style Transfer}
Most researches about style transfer focus on the combination of content and single style. In~\cite{huang2015synthesis}, initial content is transformed into oil style by pixel-level analysis. Gatys et al.~\cite{gatys2015neural} proposed a style transfer scheme based on CNN whose results are quite appealing. The model in~\cite{li2016combining} combines markov random fields with~\cite{gatys2015neural}.
In~\cite{ulyanov2016texture}, a faster feed-forward style image synthesis network is proposed.
Texture synthesis is used to transfer image style in \cite{elad2017style}. In this paper, multiple styles are combined together to synthesize appealing results with PageRank~\cite{page1999pagerank} in undirected graph~\cite{mihalcea2004textrank}.

\section{Methodology of \textsc{Line Artist}}
Our scheme includes three models: the \textit{Sketch Image Extraction} (SIE) model, the \textit{Detailed Image Synthesis} (DIS) model ,and the \textit{Adaptively Weighted Artistic Style Transfer} (AWAST) model. The overall goal of our scheme is to generate a synthesized painting $\bm{p_s}$ with artist style when receiving a freehand sketch image $\bm{\hat{s}}$. The overview of our scheme is shown in Fig.\ref{fig::overallFlow}.

\subsection{Sketch Image Extraction Model}\label{subsec::SIE}
The SIE model receives a reality image $\bm{r}$. After the L0-Smooth process, which is denoted by $l_0(\bm{\cdot})$, $\bm{r}$ is transformed into the smooth image $\bm{m}$. Then, the sketch extraction algorithm $ext(\bm{\cdot})$ receives the smooth image and produce the sketch image $s$.

\begin{equation*}
  \bm{S} = ext(l_0(\bm{R})),
  \label{equ::contributiveimg}
\end{equation*}
where $\bm{S}$ and $\bm{R}$ are sets containing $\bm{s}$ and $\bm{r}$. Then the pairs of $\bm{r}$ and $\bm{s}$ is denoted by
\begin{equation*}
  \bm{\left[R,S\right]}=\left\{(\bm{r},\bm{s})|\bm{s}=ext(l_0(\bm{r})), \bm{r}\in \bm{R}, \bm{s}\in \bm{S}\right\}.
\end{equation*}

\begin{figure}[htb]
  \centering
  \includegraphics[width = 0.8\textwidth]{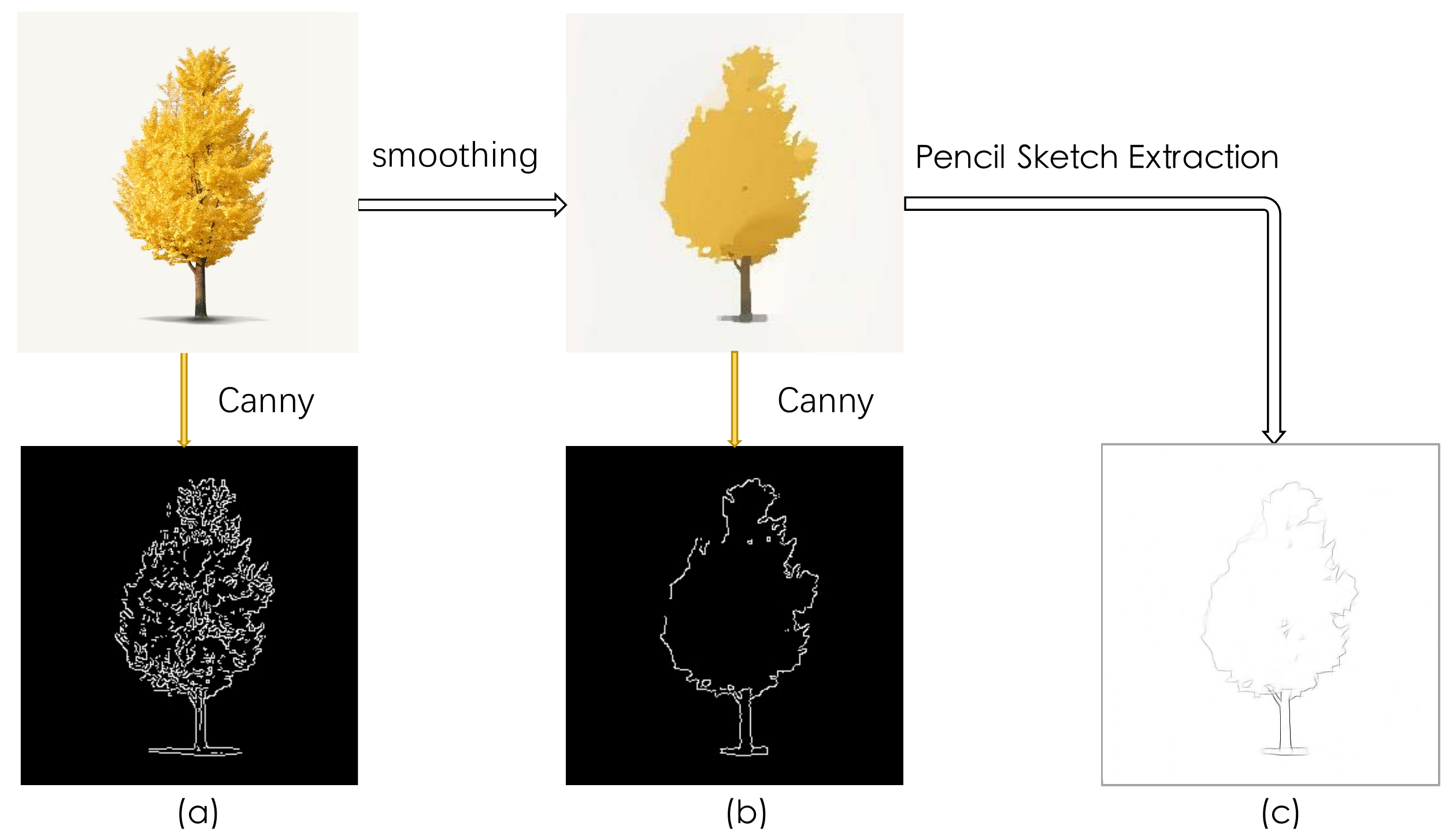}
  \caption{The procedure and results of SIE model. The upper left side is the origin image. (a) and (b) shows the differences of Canny algorithm before and after smoothing. (c) is the result of Pencil Sketch Extraction.}
  \label{fig::ceeimg}
\end{figure}

\subsubsection{Image Smoothing}\label{subsubsec::smoothing}
A challenge of sketch extraction is that there are complicated edges and patterns in the real-world images, for example, the gingko tree in Fig.~\ref{fig::ceeimg}. There are much patterns of the leaves so that the edges extracted in (a) is not similar with sketches of humans. When drawing a picture, most people cannot draw so subtle. Instead, we only draw the overall edges and simple patterns. So, the image smoothing process is necessary to make the dataset generated by SIE model more natural and similar to man-made one.

In this paper, we adopt the $L_0$-Smooth~\cite{xu2011image}, an image smoothing algorithm via $L_0$ gradient minimization. We use $M$ to denote the pixel map of the objective smooth image. For every pixel in $\bm{r}\in \bm{R}$, the gradient in $M_{i, j}$ is $\nabla M_{i, j}=(\partial_x M_{i, j}, \partial_y M_{i, j})^T$. The gradient measure $cnt(\cdot)$ in $M$ is defined as:
\begin{equation}
cnt(M) = \#\{(i, j)\mid |\partial_x M_{i, j}| + |\partial_y M_{i, j}| \neq 0\},
\label{equ::cnt}
\end{equation}
which counts the number of pixels in $M$ where $|\partial_x M_{i, j}| + |\partial_y M_{i, j}| \neq 0$.

The objective function of the optimization process is:
\begin{equation}
  \min_{M}\left\{\sum_{i, j}(M_{i, j} - R_{i, j})^2 + \lambda \cdot cnt(M) \right\}
  \label{equ::minM}
\end{equation}
In $L_0$-smooth algorithm, auxiliary variables $h_{i, j}$ and $v_{i, j}$ are introduced to calculate the minimization of Eqn.\ref{equ::minM}:

\begin{equation}
  \min_{M, h, v}\left\{\sum_{i, j}(M_{i, j} - R_{i, j})^2 + \lambda \cdot cnt(M) + \epsilon(h, v)  \right\}
  \label{equ::minMhv}
\end{equation}
with $\epsilon = \beta((\partial_xM_{i, j} - h_{i,j})^2 + (\partial_yM_{i, j}-v_{i, j}))$. Then with $M$ initialized by $\bm{r}$, the iteration will converge to the smooth image $M^*$
\subsubsection{Canny Edge Extraction}\label{subsubsec::canny}
With the smooth image $M^*$ extracted, we can extract the contributive edge image $C = canny(M^*)$ by Canny operator. Here we adopt the Canny operator with Sobel method and double threshold. The result is shown in Fig.\ref{fig::ceeimg}(b). We can see that the overall shape of the tree is extracted without unexpected noise, which is much more like human sketch. However, there are still some problems. the edges are not smooth and natural enough. When people draw, different shades will be applied to different lines and areas. However, the Canny algorithm does not produce the information of shades.

\subsubsection{Pencil Sketch Extraction}
Although a sketching-like image is already generated use Canny operator in Section~\ref{subsubsec::canny}, the line in the generated image $C$ is not natural enough compared with real freehand sketching by humans.

To solve this problem, we use the the algorithm proposed in~\cite{lu2012combining} to produce pencil sketches from real-world images. We mainly use the line drawing with strokes method, for we do not need pencil to draw the shadow.

There are some issues we cannot ingore when building the line drawing with strokes method. One is that artists always draw lines with breaks, but not long lines without any breaks. Another is that there are always crosses at the junction of two lines.
Based on these two important issues, the method of drawing lines from strokes was born. When drawing strokes at a point, we determine the direction, length, width and shade in a pixel classification and the link process based on a unified convolution framework.

\textbf{Classfication} We first transform the input image into grayscale version. Then compute the gradients of it, yielding magnitude:
	\begin{equation}
	G = ((\partial_{x}I)^{2} + (\partial_{y}I)^{2})^{\frac{1}{2}},
	\end{equation}
where I is the grayscale input, $\partial_{x}$ and $\partial_{y}$ are gradient operators in two directions, implemented by forward difference. Then we use local information to do the classification. We choose $8$(depending on the case, it may changed to $6$ or other value) directions, each is $45$ degrees apart and denote the line segment as $L_{i}$($i = 1, 2, ... , 8)$ representing the $8$ directions and the length of $L_{i}$ is $\frac{1}{30}$ of the height or width. So the responding map for a certain direction $i$ is
	\begin{equation}
	C_{i} = L_{i} * G,
	\end{equation}
where $*$ is the convolution operator. Finally, the classification is performed by choosing the maximum value among the response map in the $8$ directions. This step is written as
	\begin{equation}
		C_{i}(p) = \left\{
             \begin{array}{lr}
             G_{p}, &~if~argmin_{i}[G_{i}(p)] = i \\
             0, & otherwise
             \end{array}
\right.
\end{equation}
where $p$ refers to the pixels and $C_{i}$ is the magnitude map for direction $i$.

\textbf{Line Shaping} We generate lines also by convolution when given the map set $C_{i}$,
	\begin{equation}
	S^{'} = \sum_{i=1}^{8}(L_{i} \otimes C_{i})
	\end{equation}

Convolution aggregates nearby pixels along direction $L_{i}$, which links edge pixels that are even not connected in the original gradient map.

Using these two method, we get freehand sketches, which are more like what ordinary people draw than using the Canny method.


%
%
%
%

\subsection{Detailed Image Synthesis Model}\label{subsec::DIS}
In DIS model, We adapt the \textit{conditional Generative Adversarial Networks} in~\cite{isola2016image}, which is denoted by $cGAN(\cdot)$. This architecture fits our work because cGAN runs fast and its precision is high enough for artistic synthesis. In the training process, $cGAN$ receive the pairs of reality and sketch images, $\bm{[S, R]}$, generated in Section~\ref{subsec::SIE} to train the model. In the test process, the DIS model receives the real freehand sketch image $\hat{s}$ and then generates a detailed informative image $d$.

The loss function of CGAN is:
\begin{equation}
  \begin{split}
\mathcal{L}_{cGAN}(G, D) =& \mathbb{E}_{(s, r)\sim \bm{[S, R]}}[logD(s, r)] +\\
&\mathbb{E}_{s\sim \bm{S}, z}[log(1-D(s, G(s, z)))],
\end{split}
\end{equation}
where $G(\cdot)$ represents the generator, $D(\cdot)$ represents the discriminator.

In~\cite{isola2016image}, $G$ is set to use the U-Net~\cite{ronneberger2015u}, which is similar to the encoder-decoder architecture. However, we find that the U-Net model often leads to crash and the results become awfully meaningless noise images with strange patterns.

\begin{figure}[htb]
  \centering
  \includegraphics[width = 0.85\textwidth]{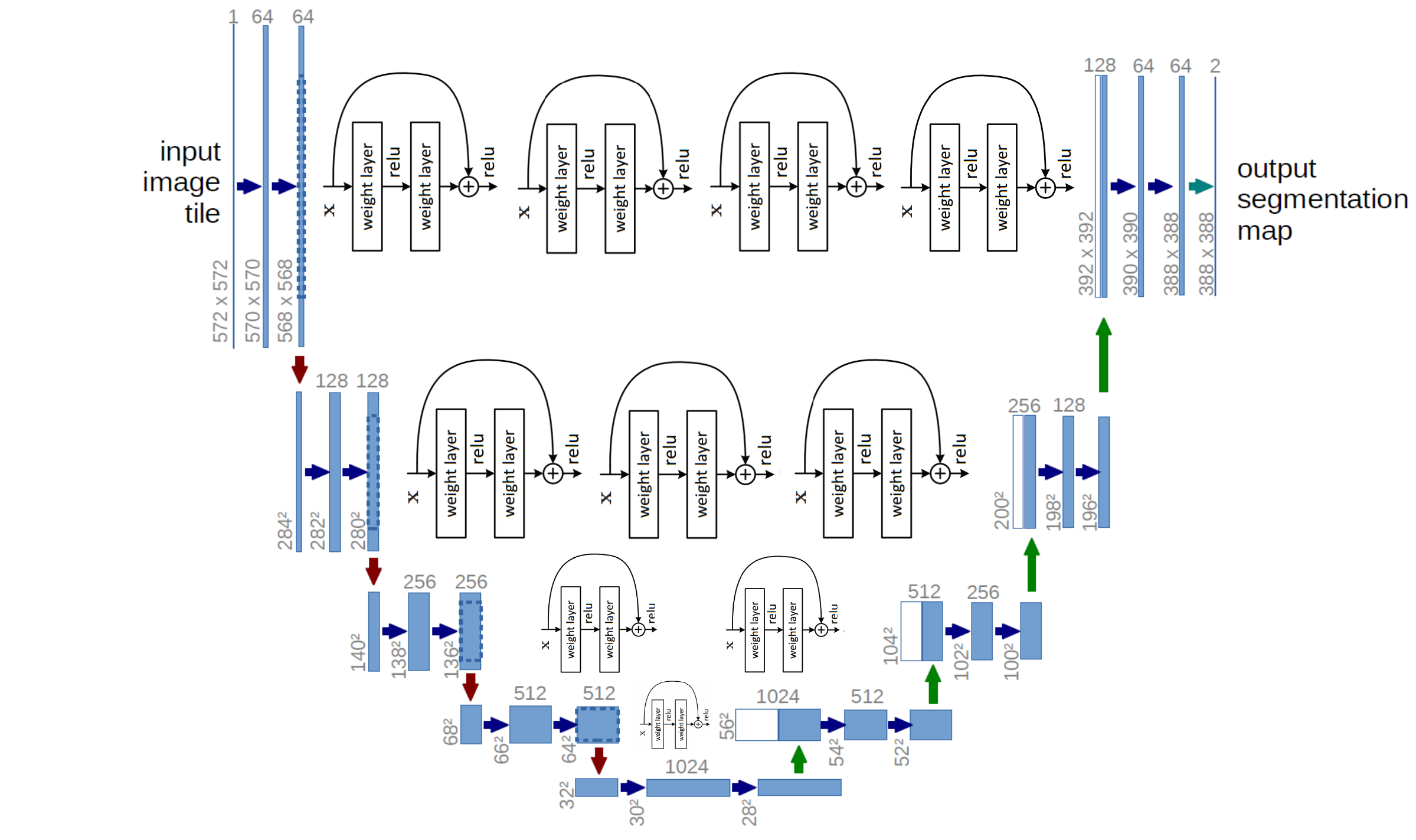}
  \caption{The architecture of Fantasy-Net. The blue arrows are convolutional operations, the red ones are pooling operations, and the green ones are transposed convolutional operations. In the middile are many residual blocks consisting of skip connections and weight layers.}
  \label{fig::fantasy}
\end{figure}

We propose a new architecture of generator named Fantasy-Net, which is shown in Fig.~\ref{fig::fantasy}. The Fantasy-Net combines the advantages of the U-Net and Residual-Net~\cite{he2016deep}. We attributes the crash problem to the simple skip connection in the U-Net. Intuitively, the simple skip connection will lead to some disorder of the convolutional layers. The residual blocks have both the features of skip connection and further encoding. So we use the residual blocks to optimize the skip connection in U-Net.

Experiments prove that these residual blocks does not increase much running time, and the crash problem is solved. The results of Fantsy-Net are also less blurring and the colors are more natural.

In order to make the results closer to reality, we adapt the objective cGAN with $L_1$ distance inspired by~\cite{pathak2016context}. By doing this, the generator is able to generate image more similar to the ground truth.

The compound objective is:
\begin{equation}
\mathcal{L} =  \mathcal{L}_{cGAN}(G, D) + \lambda \mathbb{E}_{(s, r)\sim \bm{[\bar{S}, R]}, z}[\|r-G(s, z)\|_1]
\end{equation}

Then, the objective generator by the optimization is:
\begin{equation}
 \bar{G} = \arg \min_G \max_D \mathcal{L}(G, D)
\end{equation}

with the objective $\bar{G}$ by training, we can generate the detailed image $d$ by $d = G(\hat{s})$.

%
%
%
%
%

\subsection{Adaptively Weighted Artistic Style Transfer} \label{subsec::AWAST}
The \textit{Adaptively Weighted Artistic Style Transfer} (AWAST) model recieves the detailed informative image $d$ synthesized in Section~\ref{subsec::DIS} and a given set of artistic style samples $A$. The features of the detailed image $d$ and the artistic samples $a_i\in A$ are extracted from a pretrained VGG19~\cite{simonyan2014very} net. A novel weight calculating algorithm based on PageRank is used to combine the features. Then, a random noise image $x$ was optimized to our objective image $g$, which combines both the low and high level features of $d$ and $A$.

\begin{figure}[htb]
  \centering
  \includegraphics[width = 0.98\textwidth]{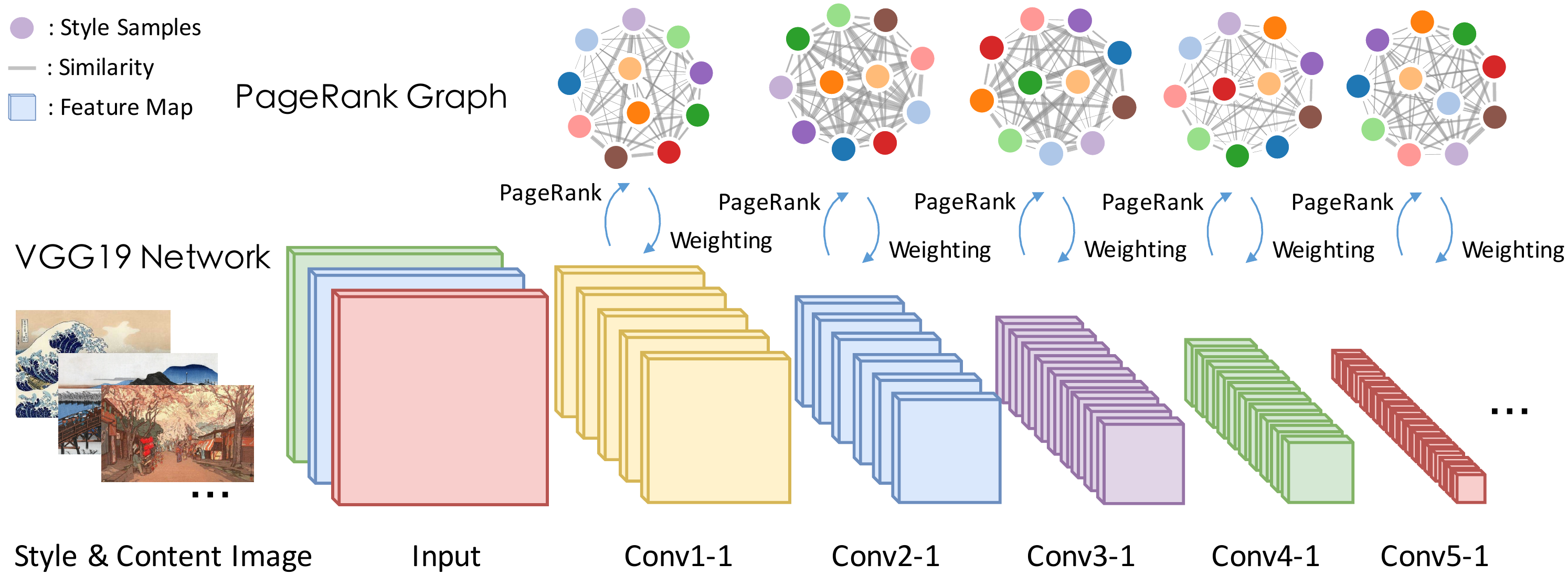}
  \caption{The AWAST model. A pretrained VGG19 network is used to downsample the style and content images. Undirected graphs are built according to the similarity between the feature maps of different style images. ASWs are calculated with PageRank algorithm and applied to weighting the features.}
  \label{fig::pagerank}
\end{figure}
The first step of AWAST is to feed $d$ and every $a_i$ into the pretrained VGG19 network inspired by the artistic style algorithm~\cite{gatys2015neural}. After this process, we obtain an activated feature map $\bm{F}_l$ on each convolutional layer $l$. Every activated feature map represents the different levels of features. In the beginning of convolutional layer, more low-level information about colors, patterns and details is extracted. And the high-level information like the distribution and shape is store in the latter convolutional layer.

We denote the set of activated feature maps for every artistic sample $a\in A$ as $\mathbb{F}_a$: $\mathbb{F}_a = \bigcup\{\bm{F}_{a,l}\}$. Similarly, we denote the extracted features of informative image $d$ as $\mathbb{F}_d = \bigcup\{\bm{F}_{d,l}\}$ and $\mathbb{F}_x = \bigcup\{\bm{F}_{x,l}\}$ for noise image $x$.

Our approach is optimizing $x$ by iteration $t$ to make sure it is similar to both $d$ and $a$ in $A$. So, we define the loss function of the optimization as:
\begin{equation}
\mathcal{L}(d, A, l, t) = \alpha \mathcal{L}_{1}(l, d, x_t) + \beta \sum_{l, a\in A}\bar{\omega}(l, a)\cdot \mathcal{L}_{2}(l, a, x_t),
\end{equation}\label{equ::overallloss}
where $\mathcal{L}_1(d, x_t)$ is the content loss between $d$ and $x_t$, $\mathcal{L}_2(a, x_t)$ is the style loss between $a$ and $x_t$. $\bar{\omega}(l, a)$ is the \textit{adaptive style weight} (ASW) of sample $a$ and layer $l$. $\alpha$ is the overall content weight and $\beta$ is the overall style weight.

The content loss $\mathcal{L}_1$ is defined as the squared error loss between $d$ and $x_t$: \begin{equation}
\mathcal{L}_1(l, d, x_t) = \frac{1}{2}\sum_{i,j}(F_{i, j, d}^{l}-F_{i, j, x_t}^{l}).
\end{equation}

The style loss is defined with the Eccentric covariance matrix (Gram) matrix $G^l$ in different layer $l$, $G^l_{i, j} = \sum_{k} F_{i, k}^l F_{j,k}^l$. Our optimization problem is converted to minimize the difference between the Gram matrix of $a$ and $x_t$. We define the style loss as
\begin{equation}
\mathcal{L}_2(l, a, x_t) = {\frac{1}{4n_lw_lh_l}\sum_{i,j}(G_{i, j, a}^{l} G_{i, j, x_t}^l)},
\end{equation}
where $\frac{1}{4n_lw_lh_l}$ is the normalization coefficient in order to make $\mathcal{L}_2$ compatible with $\mathcal{L}_1$.

Recall that different $\bm{F}_{a, l}$ can represent different features, in another word, the painting skills of different artists. For example, in Pablo Picasso's painting, the high-level feature would contain more sharp, irregular geometry and lines while the low-lever feature would include deeper colors and smooth pattern. For Claude Monet's paiting, the features are almost opposite.

To make the artistic style painting synthesized from multiple artist painting samples $a\in A$ be more delicate and appealing, we introduce the \textit{adaptive style weight} (ASW) to balance the style futures from different samples and layers. Since multiple painting samples are input to our system, various styles and skills from different artists will pile up. In order to emphasize the most common used skills, factions, and color preference, etc. and weaking some minor and unimportant skills from these artists , we use ASW to balance the importance of them.

To calculate the ASW, we build a style similarity network and use the undirected PageRank algorithm~\cite{grolmusz2012note} to calculate the ASW.
We define the difference matrix $\Delta$ between different $F_{a, l}$ by the squared error:
\begin{equation}
\Delta_{p, q}^l = \frac{1}{2}\sum_{i, j}(F_{i, j, a_p}^l - F_{i, j, a_q}^l)^2.
\end{equation}
We define the similarity matrix $\Gamma$ as:
\begin{equation}
\Gamma_{p, q}^l = \frac{ \max_{i, j}\Delta_{i, j}^l } { \Delta^l_{p, q}}.
\end{equation}
The matrix $\Gamma^l$ is symmetric, full-rank and the diagnal elements are all zero. We build a fullly connected undirected graph according to $\Gamma^l$. The weight of edges in this graph is defined as: $\mu_{i, j}^l = \Gamma^l_{i, j} / \sum_{j}\Gamma^l_{i, j}$. The weight of nodes in this graph is denoted as $PR$, representing the importance of this node in the PageRank algorithm.

The iteration formulation of undirected PageRank is:
\begin{equation}
PR_t(p_i) = \frac{1-\theta}{N_s} + \frac{\theta}{N_s} \sum_{p_j\in E(p_i)}\mu_{p_i, p_j}PR_{t-1}(p_j),
\end{equation}\label{equ::pagerank}
where $p_i$ and $p_j$ are nodes in the graph. $\theta$ is the damping factor of PageRank algorithm. $N_s = |A|$ is the number of nodes. $E(p_i)$ is the set of nodes that connecting with $p_i$

Then, the matrix $\overline{PR}$ is obtained after the PageRank algorithm converged. Then, we define the ASW with a sigmoid mapping:
\begin{equation}
\omega(l, a) = (1 + exp( 2 - \frac{4(\overline{PR} - \min(\overline{PR}))} {\max(\overline{PR})- \min(\overline{PR})}))^{-1},
\end{equation}
By normalization, the ASW $\bar{\omega}(l, a) = \frac{1}{N_lN_s}\omega(l, a)$ is obtained.

Then, by minizing the overall loss in Eqn.\ref{equ::overallloss}, the random noise image $x$ will converge to the objective image $g$.
%
%
%
%

\section{Experiments}
Experiments of our scheme and baselines are set to evaluate the qualities of the results. The experiment of AWAST is also operated to compare with the result of Gatys et al.

\textbf{Datasets:}\label{line::dataset}
We use the cats dataset~\cite{catdataset} on Kaggle and the Oxford buildings dataset~\cite{buildingdataset}.

For the cats dataset, after the preprocess of abandoning the images which mainly contains human beings, buildings or something but not cats, we choose $2620$ images to train our model.

For the Oxford Buildings Dataset. We randomly choose $4202$ building images in this dataset to carry out the training process.

For the styles samples in AWAST model, we mainly choose style images from Google for the ink painting style, the Picasso style, the Van gogh style, the watercolor style and Ukiyo-e style. For the Onmyoji style, we choose style images from a popular mobile game named Onmyoji.

\textbf{Environment:}
All the experiments are conducted on a PC device (Intel(R) Core(TM) i7-6900K 3.2GHz, 16GB memory, NVIDIA GeForce(R) GTX 1080 Ti) and are implemented in Python 3.6.2.

\subsection{Training}
Because our model is the combination of the three models mentioned above, we now describe the training process in three main steps corresponding to the three models.

In the SIE model, for smoothing, we set the parameters lamda to be $0.02$, kappa to be $1.2$. And for pencil sketching, we set the length of convolution line to be $7$, the width of the stroke to be $0.1$, the number of directions to be $10$ for the cat dataset. For the building dataset, we set the length of convolution line to be $7$, the width of the stroke to be $0.1$, the number of directions to be $10$. Because for some image in the two datasets, there are some pictures too complex for freehand drawing, we deleted these complex pictures.

The DIS model receives the training datasets from SIE, which are pairs of sketch images and the reality images. These images are resized to $256\times 256$ to accelerate our training speed. Then, we set the learning rate $lr = 0.0002$, the generator use the unet 256 network. The initialization of network is set to use xavier initializer. Both the input and output channels are set to be $3$.

The AWAST model receives the $256\times 256$-size images synthesized by the DIS model and the style image samples illustrated in Section~\ref{line::dataset}. The parameters $\alpha$ and $\beta$ in Eqn.\ref{equ::overallloss} are set as $8$ and $500$. The dumping factor of PageRank $\theta$ in Eqn.~\ref{equ::pagerank} is set to be $0.85$ and the convergent accuracy of PageRank is $0.0001$. The learning rate is $1$ and parameters $\beta_1$ and $\beta_2$ for Adam optimizer are $0.9$ and $0.999$. The iteration of optimization is set to be $2000$.

\subsection{Baselines and Analysis}\label{subsec::baselines}

\subsubsection{Baselines}
\begin{figure}[!ht]
  \centering
  \begin{subfigure}[b]{0.18\textwidth}
    \includegraphics[width=\textwidth]{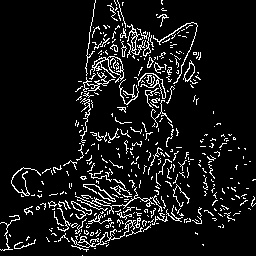}
    \caption{Canny edge}
    \label{fig::base11}
  \end{subfigure}
  ~
  \begin{subfigure}[b]{0.18\textwidth}
    \includegraphics[width=\textwidth]{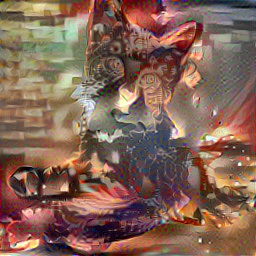}
    \caption{Onmyoji}
    \label{fig::base12}
  \end{subfigure}
  \begin{subfigure}[b]{0.18\textwidth}
    \includegraphics[width=\textwidth]{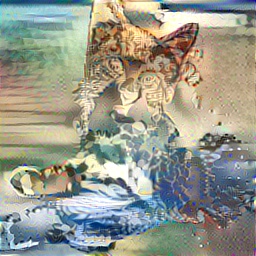}
    \caption{Ukiyo-e}
    \label{fig::base13}
  \end{subfigure}
  \begin{subfigure}[b]{0.18\textwidth}
    \includegraphics[width=\textwidth]{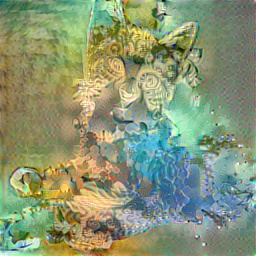}
    \caption{Van Gogh}
    \label{fig::base14}
  \end{subfigure}
  \begin{subfigure}[b]{0.18\textwidth}
    \includegraphics[width=\textwidth]{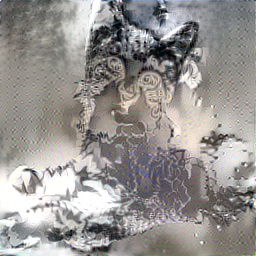}
    \caption{Chinese}
    \label{fig::base15}
  \end{subfigure}
  \vspace{-2mm}
  \caption{Baseline 1: Extract edge with canny algorithm and feed AWAST directly}
  \label{fig::cannyonly}
  \vspace{4mm}
  \begin{subfigure}[b]{0.18\textwidth}
    \includegraphics[width=\textwidth]{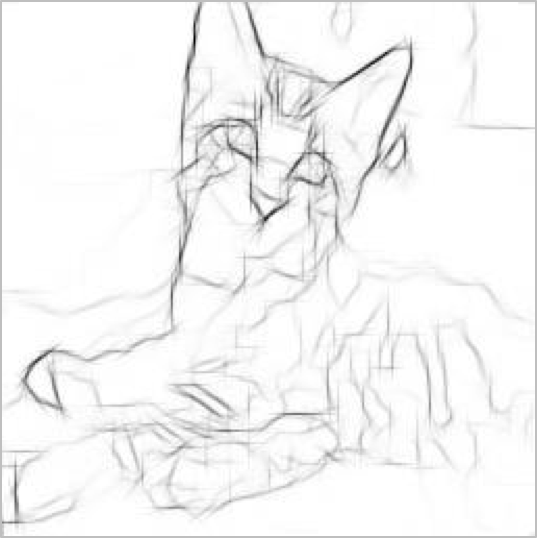}
    \caption{Sketch}
    \label{fig::base21}
  \end{subfigure}
  ~
  \begin{subfigure}[b]{0.18\textwidth}
    \includegraphics[width=\textwidth]{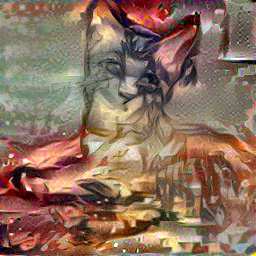}
    \caption{Onmyoji}
    \label{fig::base22}
  \end{subfigure}
  \begin{subfigure}[b]{0.18\textwidth}
    \includegraphics[width=\textwidth]{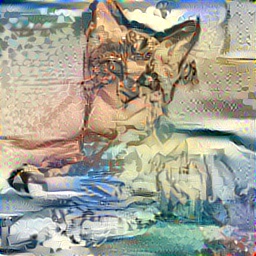}
    \caption{Ukiyo-e}
    \label{fig::base23}
  \end{subfigure}
  \begin{subfigure}[b]{0.18\textwidth}
    \includegraphics[width=\textwidth]{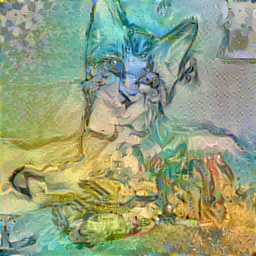}
    \caption{Van Gogh}
    \label{fig::base24}
  \end{subfigure}
  \begin{subfigure}[b]{0.18\textwidth}
    \includegraphics[width=\textwidth]{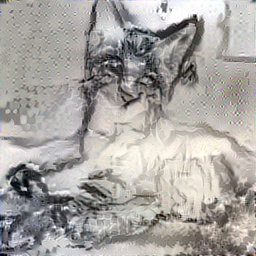}
    \caption{Chinese}
    \label{fig::base25}
  \end{subfigure}
  \vspace{-2mm}
  \caption{Baseline 2: Smoothing, extracting pencil sketh, and feed AWAST directly}
  \label{fig::sketchonly}
  \vspace{4mm}
  \begin{subfigure}[b]{0.18\textwidth}
    \includegraphics[width=\textwidth]{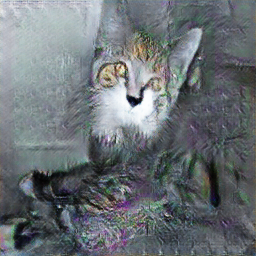}
    \caption{Detailed img}
    \label{fig::base31}
  \end{subfigure}
  ~
  \begin{subfigure}[b]{0.18\textwidth}
    \includegraphics[width=\textwidth]{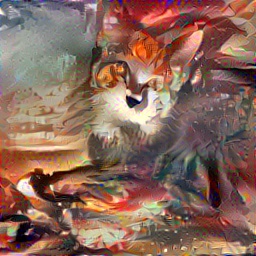}
    \caption{Onmyoji}
    \label{fig::base32}
  \end{subfigure}
  \begin{subfigure}[b]{0.18\textwidth}
    \includegraphics[width=\textwidth]{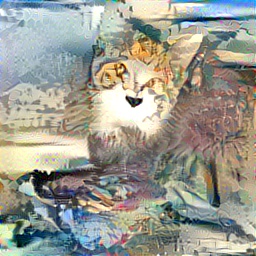}
    \caption{Ukiyo-e}
    \label{fig::base33}
  \end{subfigure}
  \begin{subfigure}[b]{0.18\textwidth}
    \includegraphics[width=\textwidth]{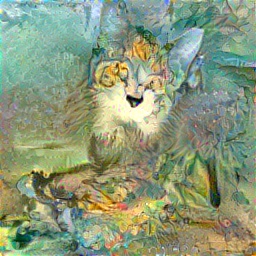}
    \caption{Van Gogh}
    \label{fig::base34}
  \end{subfigure}
  \begin{subfigure}[b]{0.18\textwidth}
    \includegraphics[width=\textwidth]{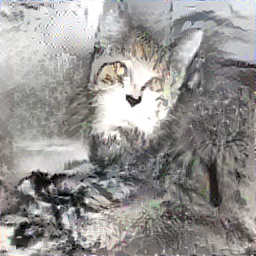}
    \caption{Chinese}
    \label{fig::base35}
  \end{subfigure}
  \vspace{-2mm}
  \caption{Baseline 3: Smoothing, Canny, DIS, and feeding AWAST with the detailed image}
  \label{fig::cannyplus}
  \vspace{4mm}
  \begin{subfigure}[b]{0.18\textwidth}
    \includegraphics[width=\textwidth]{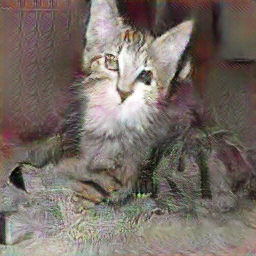}
    \caption{Detailed Img}
    \label{fig::base41}
  \end{subfigure}
  ~
  \begin{subfigure}[b]{0.18\textwidth}
    \includegraphics[width=\textwidth]{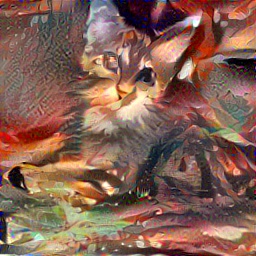}
    \caption{Onmyoji}
    \label{fig::base42}
  \end{subfigure}
  \begin{subfigure}[b]{0.18\textwidth}
    \includegraphics[width=\textwidth]{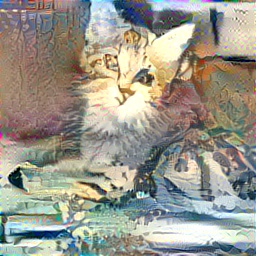}
    \caption{Ukiyo-e}
    \label{fig::base43}
  \end{subfigure}
  \begin{subfigure}[b]{0.18\textwidth}
    \includegraphics[width=\textwidth]{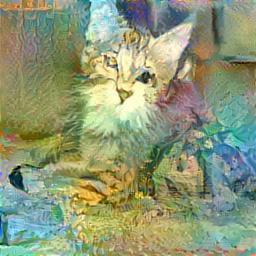}
  \caption{Van Gogh}
    \label{fig::base44}
  \end{subfigure}
  \begin{subfigure}[b]{0.18\textwidth}
    \includegraphics[width=\textwidth]{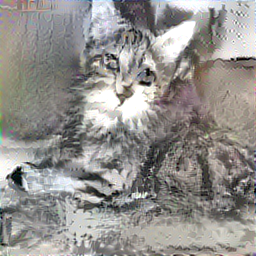}
    \caption{Chinese}
    \label{fig::base45}
  \end{subfigure}
  \vspace{-2mm}
  \caption{Our scheme: Smoothing, extracting of pencil sketch, DIS and feeding AWAST with the detailed image}
  \label{fig::sketchplus}
\end{figure}

We set some baselines to show our method of assembling the model works best.
\begin{itemize}
\item \textbf{Baseline 1: } Real-world image $\rightarrow$ Canny edges $\rightarrow$ Train the model $\rightarrow$ Artistic image. Result is shown in Fig.\ref{fig::cannyonly}.
\item \textbf{Baseline 2: } Real-world image $\rightarrow$ Smoothing image $\rightarrow$ Sketch image $\rightarrow$ Train the model $\rightarrow$ Artistic image. Result is shown in Fig.\ref{fig::sketchonly}.
\item \textbf{Baseline 3: } Real-world image $\rightarrow$ Smoothing image $\rightarrow$ Canny image $\rightarrow$ Train the model $\rightarrow$ Detailed image $\rightarrow$ Artistic image. Result is shown in Fig.\ref{fig::cannyplus}.
\item \textbf{Our scheme: } Real-world image $\rightarrow$ Smoothing image $\rightarrow$ Pencil sketch $\rightarrow$ Train the model $\rightarrow$ Detailed image $\rightarrow$ Artistic image. Result is shown in Fig.\ref{fig::sketchplus}.
\end{itemize}

\subsubsection{Analysis of baselines}
As shown in Fig.\ref{fig::cannyonly}, the synthesized artistic image is quite blurring. There are many noisy points that can't converge to the optimizing object. And the features of the content are not extracted successfully. So there are some areas just copy the pattern of the style samples. For example, in Fig.\ref{fig::base13}, near the cat's feet, the pattern is just like the tide which comes from the style samples of Ukiyo-e style. This is because the information in the edge picture is not enough to push the AWAST synthesize the objetive result, so as to people's freehand sketch. So, the information generating process, DIS, is needed.

In Fig.\ref{fig::sketchonly}, we can see that by smoothing and pencil sketch extraction, the results are much better. The problems of pattern copying like Baseline $1$ are improved and the shape of the cat looks more clear. At the same time, the style image is not so blurring as Baseline $1$. These improvements suggest that the smoothing and pencil sketch extraction are quite helpful to extract the key information in the reality image while weak the noise. By comparing Fig.\ref{fig::base11} and Fig.\ref{fig::base21}, the sketch in Baseline $2$ is more like human freehand sketch on the continuity and directions.

Fig.\ref{fig::cannyplus} represents the final results of Baseline $3$, which includes smoothing, canny, DIS, and AWAST. Compared to Baseline $1$, the results are much more beautiful. The shape of the cat is more clear and the problem of pattern copying is solved. The shade of the styles is also more natural. The details are also more clear. In Fig.\ref{fig::base35}, for example, the eyes of the cat become yellow which is different from the color of its body. This means the detail of the cat's eyes is noticed and emphasize by the AWAST model.

Our final scheme is shown in Fig.\ref{fig::sketchplus}， including smoothing, pencil sketch extraction, DIS, and AWAST. The performance improves a lot comparing to Baseline $3$.
For example, in Fig.\ref{fig::base45}, eyes of the cat are much more vivid and the patterns of fur and beard on its head and body are detailed.

On Fig.\ref{fig::base42}, the style transfer is active and does not break the shape of the cat while on Fig.\ref{fig::base32}, the style mix the cat's body and the environment.

\subsubsection{Analysis of AWAST}
\begin{figure}[!ht]
  \centering
  \begin{subfigure}[b]{0.31\textwidth}
    \includegraphics[width=\textwidth]{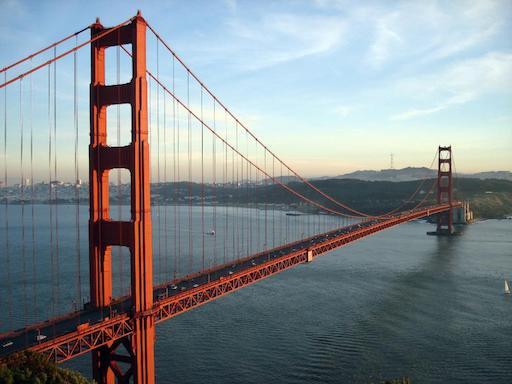}
    \caption{Reality (Golden Gate)}
    \label{fig::goldreal}
  \end{subfigure}
  \begin{subfigure}[b]{0.31\textwidth}
    \includegraphics[width=\textwidth]{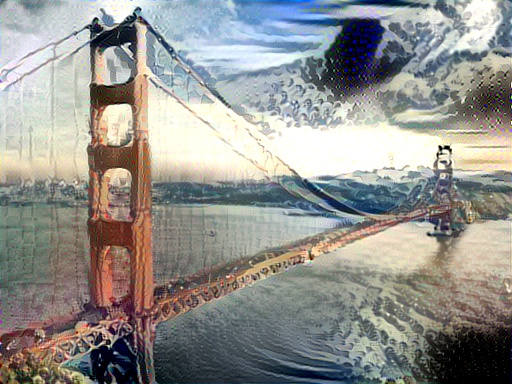}
    \caption{Result of Gatys et al.}
    \label{fig::goldgatys}
  \end{subfigure}
  \begin{subfigure}[b]{0.31\textwidth}
    \includegraphics[width=\textwidth]{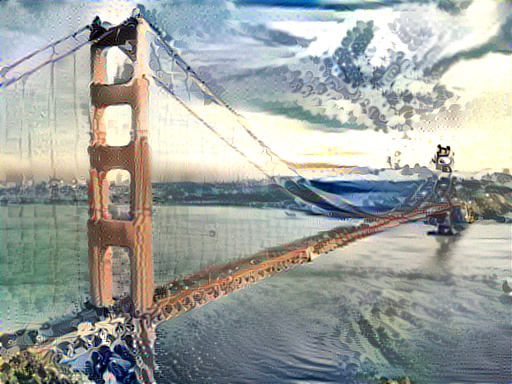}
    \caption{Result of Ours}
    \label{fig::goldours}
  \end{subfigure}
  \caption{Performance analysis of AWAST compared to Gatys et al.}
  \label{fig::goldbridge}
  \vspace{3mm}
  \begin{subfigure}[b]{0.8\textwidth}
    \includegraphics[width=\textwidth]{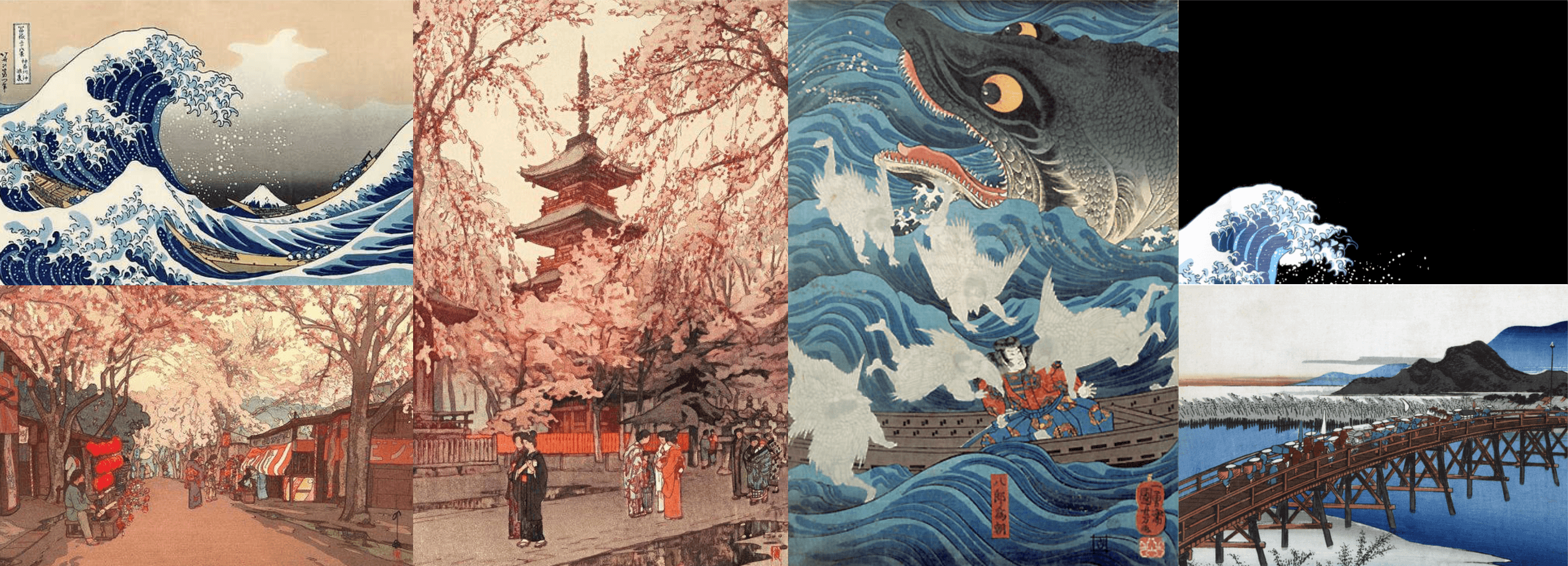}
  \end{subfigure}
  \caption{Images of Ukiyo-e style to analyze AWAST. There are lots of black areas in one of the images.}
  \label{fig::fshstyle}
\end{figure}

We use $6$ images in Fig~\ref{fig::fshstyle} of Ukiyo-e style to analyze the performance of AWAST. However, one image of them is coverd by black color. The experiment result when iterations $iter=2500$ is shown in Fig.\ref{fig::goldbridge}. The result of Gatys et al. in Fig.\ref{fig::goldgatys} appears to have many black areas, which is attributed to taking the patterns of black. This is because the blending method of them is simply taking the average of multiple features. Our method in Fig.\ref{fig::goldours} is steadier and more natural, because ASW based on PageRank algorithm weaks the weight of the unusual styles like black images and emphasizes the right styles in the other $5$ images.

%

\subsection{Test of man-made sketch}
\begin{figure}[!ht]
\begin{subfigure}[b]{0.16\textwidth}
  \includegraphics[width=\textwidth]{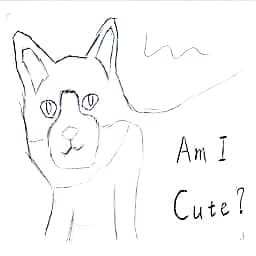}
  \caption{Sketch}
  \label{fig::freehand}
\end{subfigure}
\begin{subfigure}[b]{0.16\textwidth}
  \includegraphics[width=\textwidth]{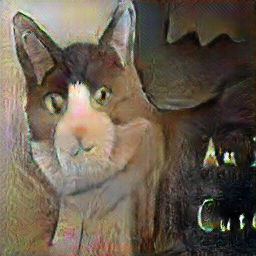}
  \caption{Detailed}
  \label{fig::realdetailed}
\end{subfigure}
\begin{subfigure}[b]{0.16\textwidth}
  \includegraphics[width=\textwidth]{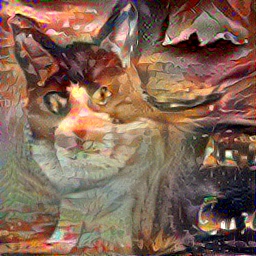}
  \caption{Onmyoji}
  \label{fig::realyys}
\end{subfigure}
\begin{subfigure}[b]{0.16\textwidth}
  \includegraphics[width=\textwidth]{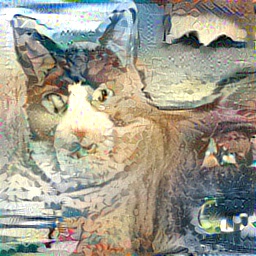}
  \caption{Ukiyo-e}
  \label{fig::reslfsh}
\end{subfigure}
\begin{subfigure}[b]{0.16\textwidth}
  \includegraphics[width=\textwidth]{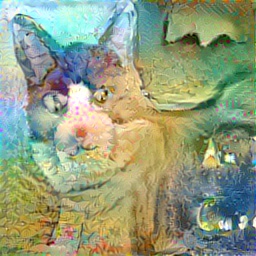}
\caption{Van Gogh}
  \label{fig::realfg}
\end{subfigure}
\begin{subfigure}[b]{0.16\textwidth}
  \includegraphics[width=\textwidth]{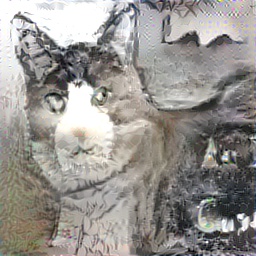}
  \caption{Chinese}
  \label{fig::realchinese}
\end{subfigure}
\caption{Test result of real freehand sketch of cat. (a) is the man-made sketch. (b) is the detailed image synthesized by DIS. (c)-(f) are artistic paintings.}
\label{fig::realtest}
\end{figure}

Most of the experiments above is based on the test set generated by SIE model. However, our general objective is to transform a real man-made sketch to an artistic painting. So we take an $A4$ paper and a pencil and draw a cat sketch by hand, which is shown in Fig.\ref{fig::freehand}. This is quite easy actually. We simply take a photo with our mobile-phone and then feed it into the DIS and the AWAST model. The result is shown in Fig.\ref{fig::realtest}.

\vspace{-8mm}
\section{Conclusions}
To be an artist is always many people's dream. In this paper, we propose \textsc{Line Artist} to synthesize appealing paintings with freehand sketch. To achieve this goal, we propose the SIE model to smooth the images and extract the sketch to build new datasets. The DIS model based on GAN allow the sketch drew by users to generate more details and looks natural. The AWAST model propose an novel algorithm based on PageRank to adaptively combining styles from multiple painting samples. Our results are prove to be vivid, artistic, and adapted to different styles stably. There are also some issues need to be improved. In the DIS and the AWAST model, the synthesis is not real-time. The quality of detailed images synthesized by DIS is also not so satisfactory because DIS is a process to generate more information with limited conditions. However, \textsc{Line Artist} still has the potential to become an powerful entertainment APP and assistant of artists.

\bibliographystyle{splncs}
\bibliography{egbib}

\begin{thebibliography}{10}

\bibitem{gatys2015neural}
Gatys, L.A., Ecker, A.S., Bethge, M.:
\newblock A neural algorithm of artistic style.
\newblock arXiv preprint arXiv:1508.06576 (2015)

\bibitem{page1999pagerank}
Page, L., Brin, S., Motwani, R., Winograd, T.:
\newblock The pagerank citation ranking: Bringing order to the web.
\newblock Technical report, Stanford InfoLab (1999)

\bibitem{xu2011image}
Xu, L., Lu, C., Xu, Y., Jia, J.:
\newblock Image smoothing via l0 gradient minimization.
\newblock In: ACM Transactions on Graphics (TOG). Volume~30., ACM (2011)  174

\bibitem{lu2012combining}
Lu, C., Xu, L., Jia, J.:
\newblock Combining sketch and tone for pencil drawing production.
\newblock In: Proceedings of the Symposium on Non-Photorealistic Animation and
  Rendering, Eurographics Association (2012)  65--73

\bibitem{isola2016image}
Isola, P., Zhu, J.Y., Zhou, T., Efros, A.A.:
\newblock Image-to-image translation with conditional adversarial networks.
\newblock arXiv preprint arXiv:1611.07004 (2016)

\bibitem{simonyan2014very}
Simonyan, K., Zisserman, A.:
\newblock Very deep convolutional networks for large-scale image recognition.
\newblock arXiv preprint arXiv:1409.1556 (2014)

\bibitem{gonzalez2016improved}
Gonzalez, C.I., Melin, P., Castro, J.R., Mendoza, O., Castillo, O.:
\newblock An improved sobel edge detection method based on generalized type-2
  fuzzy logic.
\newblock Soft Computing \textbf{20}(2) (2016)  773--784

\bibitem{kumar2017fractional}
Kumar, S., Saxena, R., Singh, K.:
\newblock Fractional fourier transform and fractional-order calculus-based
  image edge detection.
\newblock Circuits, Systems, and Signal Processing \textbf{36}(4) (2017)
  1493--1513

\bibitem{gao2010improved}
Gao, W., Zhang, X., Yang, L., Liu, H.:
\newblock An improved sobel edge detection.
\newblock In: Computer Science and Information Technology (ICCSIT), 2010 3rd
  IEEE International Conference on. Volume~5., IEEE (2010)  67--71

\bibitem{Dollar_2013_ICCV}
Dollar, P., Zitnick, C.L.:
\newblock Structured forests for fast edge detection.
\newblock In: The IEEE International Conference on Computer Vision (ICCV).
  (December 2013)

\bibitem{xie15hed}
"Xie, S., Tu, Z.:
\newblock Holistically-nested edge detection.
\newblock In: Proceedings of IEEE International Conference on Computer Vision.
  (2015)

\bibitem{wei2017road}
Wei, Y., Wang, Z., Xu, M.:
\newblock Road structure refined cnn for road extraction in aerial image.
\newblock IEEE Geoscience and Remote Sensing Letters \textbf{14}(5) (2017)
  709--713

\bibitem{zhang2015face}
Zhang, S., Gao, X., Wang, N., Li, J., Zhang, M.:
\newblock Face sketch synthesis via sparse representation-based greedy search.
\newblock IEEE transactions on image processing \textbf{24}(8) (2015)
  2466--2477

\bibitem{zhang2015segment}
Zhang, F., Dai, L., Xiang, S., Zhang, X.:
\newblock Segment graph based image filtering: fast structure-preserving
  smoothing.
\newblock In: Proceedings of the IEEE International Conference on Computer
  Vision. (2015)  361--369

\bibitem{sangkloy2016scribbler}
Sangkloy, P., Lu, J., Fang, C., Yu, F., Hays, J.:
\newblock Scribbler: Controlling deep image synthesis with sketch and color.
\newblock arXiv preprint arXiv:1612.00835 (2016)

\bibitem{odena2016conditional}
Odena, A., Olah, C., Shlens, J.:
\newblock Conditional image synthesis with auxiliary classifier gans.
\newblock arXiv preprint arXiv:1610.09585 (2016)

\bibitem{chen2017photographic}
Chen, Q., Koltun, V.:
\newblock Photographic image synthesis with cascaded refinement networks.
\newblock arXiv preprint arXiv:1707.09405 (2017)

\bibitem{ho2016laplacian}
Ho~Lee, J., Choi, I., Kim, M.H.:
\newblock Laplacian patch-based image synthesis.
\newblock In: Proceedings of the IEEE Conference on Computer Vision and Pattern
  Recognition. (2016)  2727--2735

\bibitem{pathak2016context}
Pathak, D., Krahenbuhl, P., Donahue, J., Darrell, T., Efros, A.A.:
\newblock Context encoders: Feature learning by inpainting.
\newblock In: Proceedings of the IEEE Conference on Computer Vision and Pattern
  Recognition. (2016)  2536--2544

\bibitem{huang2015synthesis}
Huang, F., Wu, B.H., Huang, B.R.:
\newblock Synthesis of oil-style paintings.
\newblock In: Pacific-Rim Symposium on Image and Video Technology, Springer
  (2015)  15--26

\bibitem{li2016combining}
Li, C., Wand, M.:
\newblock Combining markov random fields and convolutional neural networks for
  image synthesis.
\newblock In: Proceedings of the IEEE Conference on Computer Vision and Pattern
  Recognition. (2016)  2479--2486

\bibitem{ulyanov2016texture}
Ulyanov, D., Lebedev, V., Vedaldi, A., Lempitsky, V.S.:
\newblock Texture networks: Feed-forward synthesis of textures and stylized
  images.
\newblock In: ICML. (2016)  1349--1357

\bibitem{elad2017style}
Elad, M., Milanfar, P.:
\newblock Style transfer via texture synthesis.
\newblock IEEE Transactions on Image Processing \textbf{26}(5) (2017)
  2338--2351

\bibitem{mihalcea2004textrank}
Mihalcea, R., Tarau, P.:
\newblock Textrank: Bringing order into text.
\newblock In: Proceedings of the 2004 conference on empirical methods in
  natural language processing. (2004)

\bibitem{ronneberger2015u}
Ronneberger, O., Fischer, P., Brox, T.:
\newblock U-net: Convolutional networks for biomedical image segmentation.
\newblock In: International Conference on Medical Image Computing and
  Computer-Assisted Intervention, Springer (2015)  234--241

\bibitem{he2016deep}
He, K., Zhang, X., Ren, S., Sun, J.:
\newblock Deep residual learning for image recognition.
\newblock In: Proceedings of the IEEE conference on computer vision and pattern
  recognition. (2016)  770--778

\bibitem{grolmusz2012note}
Grolmusz, V.:
\newblock A note on the pagerank of undirected graphs.
\newblock arXiv preprint arXiv:1205.1960 (2012)

\bibitem{catdataset}
Research, M.:
\newblock Dogs vs. cats dataset.
\newblock \url{https://www.kaggle.com/c/dogs-vs-cats} Accessed March 11, 2018.

\bibitem{buildingdataset}
James~Philbin, R.A., Zisserman, A.:
\newblock The oxford buildings dataset.
\newblock \url{http://www.robots.ox.ac.uk/~vgg/data/oxbuildings/} Accessed
  March 11, 2018.

\end{thebibliography}

\end{document}